\documentclass[letterpaper, 10 pt, conference]{ieeeconf}
\IEEEoverridecommandlockouts
\usepackage{makecell}
\usepackage{multirow}
\usepackage{booktabs}
\usepackage{threeparttable}
\usepackage{colortbl}
\usepackage{hhline}
\usepackage{color}
\usepackage{amsmath,amsfonts}
\usepackage[table,xcdraw]{xcolor}

\usepackage{algorithm}
\usepackage[english]{babel}
\usepackage{vcell}
\usepackage{tablefootnote}
\usepackage[utf8]{inputenc}
\usepackage[T1]{fontenc}

\usepackage{cuted}
\usepackage{caption}

\usepackage{algorithmic}

\makeatletter
\newcommand{\removelatexerror}{\let\@latex@error\@gobble}
\makeatother

\usepackage{array}

\usepackage[caption=false,font=normalsize,labelfont=sf,textfont=sf]{subfig}
\usepackage{textcomp}
\usepackage{stfloats}
\usepackage{verbatim}
\usepackage{graphicx}
\usepackage{epsfig}
\usepackage{hyperref}
\DeclareUnicodeCharacter{2061}{}
\usepackage[backend=bibtex,style=ieee,sorting=none,isbn=false,url=false,doi=false]{biblatex}
\title{\LARGE \bf
Theoretical Modeling and Bio-inspired Trajectory Optimization of A Multiple-locomotion Origami Robot
}

\author{Keqi Zhu$^{1, *}$, Haotian Guo$^{1, *}$, Wei Yu$^1$, Hassen Nigatu$^{1, 2}$, Tong Li$^3$, Huixu Dong$^{1,2,\dag} $
\thanks{ $*$ Keqi Zhu and Haotian Guo contributed equally to this work}
\thanks{ 1 Robot Perception and Grasp Laboratory(Grasp Lab), Zhejiang University, Hangzhou 310058, China. Corresponding author Huixu Dong, e-mail: huixudong@zju.edu.cn.}
\thanks{ 2 Robotics Institute of Zhejiang University(robotics research center of yuyao).}
\thanks{ 3 Department of Sports Science, Zhejiang University, 310058, Hangzhou, China}
\thanks{This work was supported by the startup fund of "New Hundredd-Talent Program" of Zhejiang University under Grant 109100+194422R3/008 and Grant 109100$*$1942222R3/009 from Huixu Dong.}
}

\begin{document}
\maketitle
\begin{abstract}
Recent research on mobile robots has focused on increasing their adaptability to unpredictable and unstructured environments using soft materials and structures. However, the determination of key design parameters and control over these compliant robots are predominantly iterated through experiments, lacking a solid theoretical foundation. To improve their efficiency, this paper aims to provide mathematics modeling over two locomotion, crawling and swimming. Specifically, a dynamic model is first devised to reveal the influence of the contact surfaces' frictional coefficients on displacements in different motion phases. Besides, a swimming kinematics model is provided using coordinate transformation, based on which, we further develop an algorithm that systematically plans human-like swimming gaits, with maximum thrust obtained. The proposed algorithm is highly generalizable and has the potential to be applied in other soft robots with multiple joints. Simulation experiments have been conducted to illustrate the effectiveness of the proposed modeling. 

\end{abstract}


\section{Introduction}

With the burgeoning advancements in materials science and structural science, soft robotics has emerged as a field of significant interest and extensive research, notably when contrasted with traditional robots, which are often characterized by their intricate and sizable electromechanical systems\cite{ref2}. These contemporary soft robots are distinguished by their apparent benefits, namely their lightweight, compact construction, and the cost-effectiveness of their mass fabrication capabilities. Nevertheless, the efficacy of these robots within various environments is intrinsically linked to the optimization of their structural design parameters and proper control strategies.

The design of the soft robot plays a decisive role in determining the upper limit of the robot's performance. Notably, Li et al. designed a soft robot through both systematic experiments and theoretical analyses, which can be actuated successfully in a field test in the Mariana Trench down to a depth of 10,900 meters\cite{tiefeng}. However, many studies focus on the bio-inspiration analysis and function realization of soft robots, and few of them further consider the optimization of structural parameters to improve motion efficiency \cite{coconut,octopus}. Using the black box method, Morzadec et al. focused on shape optimization, which can skip the complex modeling process but is time-consuming\cite{blackbox}. Vikas et al. proposed a cost function to optimize the key parameters, without the modeling\cite{designfriction}. Furthermore, a subset of research has concentrated on the application of topological approaches\cite{topology} and differentiable simulation techniques\cite{diff} for the optimization of design parameters. The modeling processes employed by these methods are inherently more complex, and their transferability to other robots appears to be somewhat limited. Currently, despite the rapid development of soft robot design, there still exists a lack of how to efficiently combine the structural characteristics to improve robots' performance.  

\begin{figure}[t]
\centerline{\includegraphics[width=0.8\columnwidth]{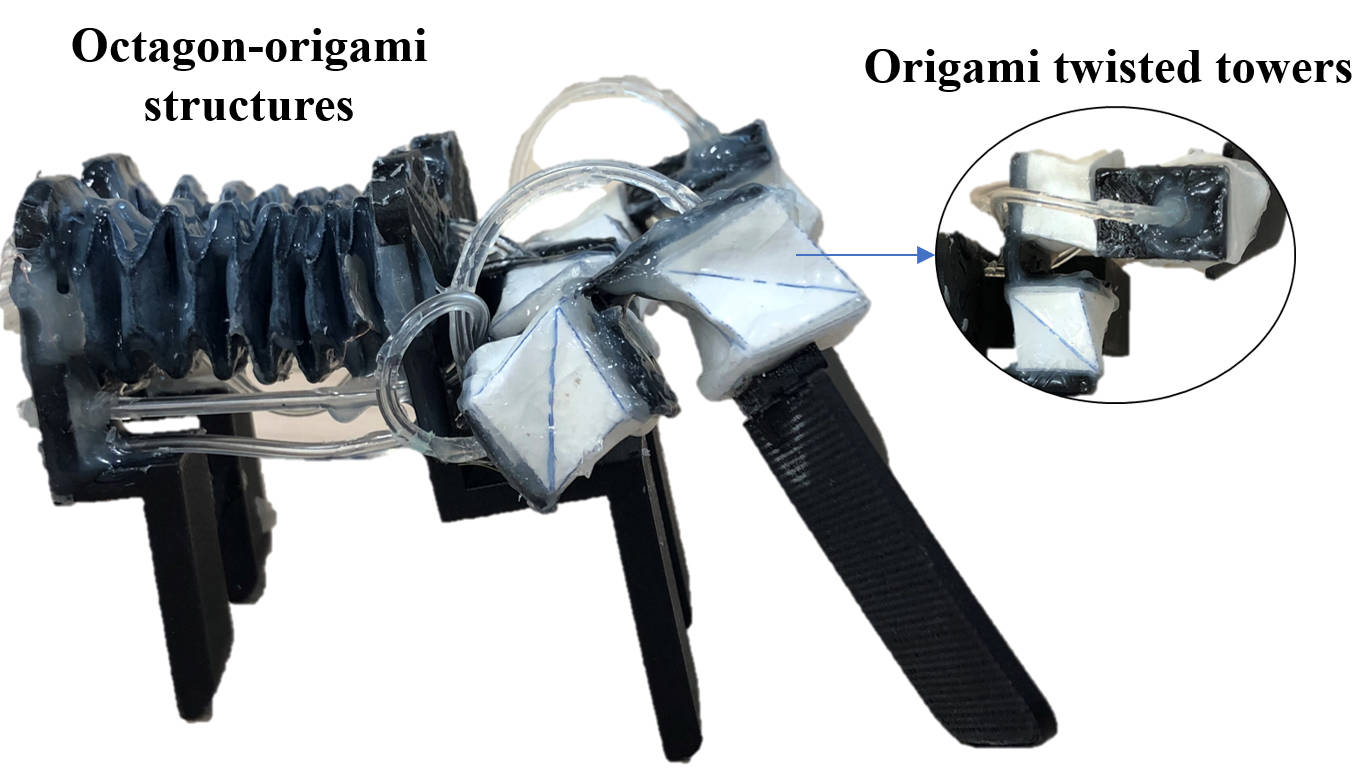}}
\vspace{-2mm}
\caption{\small{The diagram of the origami bio-inspired robot, retrieved from \cite{dong}.}}
\label{Fig:1}
\vspace{-6mm}
\end{figure}
The performance of the soft robots relies not only on their structural design but heavily on their control strategy or planned gaits. Usually, this is obtained via bio-inspiration analysis, due to the compliant and complex nature of these soft robots. For example, Tang et al. developed a frog's webbed feet-inspired swimming robot, with the gait obtained experimentally \cite{frog}. This is true of Yang's research, where they determined the gaits through repetitive trials \cite{soro}. These experiment-based design optimizations are time-consuming and often fail to obtain the optimal gaits that make full use of the provided robot design. Notable works include Li \cite{2023soft} and Kim \cite{tro}. In the former, reinforcement learning is applied to enable a snake robot to master two swimming patterns, either with the fastest swimming speed or with the least power consumption. However, a simulation platform is needed to collect massive data for the training. In the latter, a systematical dynamics model of a multimodal underwater robot is established, enabling the robots to delicately walk, swim, and actively control buoyancy. However, their methods are too complex to be generalized to more soft robots. Providing an efficient method that is compatible with a wider variety of soft robots is challenging.

Origami-based soft robots, due to the constraints introduced with the origami structures, facilitate both the design and control of soft robots for various applications and thus attract researchers' attention \cite{ref1,ref4}. 
Without loss of generality, the purpose of this study is to provide a mathematical analysis of the physical model of such a robot in terms of crawling and swimming locomotion, taking the multi-locomotion origami robot in \cite{dong} as an example (as seen in Fig.\ref{Fig:1}), to present design optimization and a generalizable gait planning algorithm, facilitating more efficient crawling and swimming motions. The proposed origami robot has octagon-origami structures as the body designed for crawling and the twisted tower origami-based arm for swimming. Especially, We qualitatively explore the effects of the proposed robot's design parameters on multiple-locomotion, which provides some insights for constructing a multiple-locomotion origami robot, and an algorithm is proposed, providing hints for how human-like optimized swimming gaits reasoning. We analytically test the proposed robot with an excellent qualitative agreement with the locomotion performance and the theoretical models. The contributions of the paper are summarized below
\begin{enumerate}
    \item We construct theoretical modeling regarding the geometric configurations of robots consisting of two origami patterns, octagon origami and twisted tower origami, for crawling and swimming. 
    \item We integrate the periodic dynamic model of the robots with the design parameters and motion variables, enabling the quantification of the influence of surface coefficient on crawling motion and providing hints for soft robot design.
    \item An heuristic-based algorithm has been developed to optimize swimming gaits and has the potential to be generalized to a variety of multi-joint articulated structures consisting of soft actuation elements. 
\end{enumerate}

\section{Methodology}
Taking the multi-locomotion origami robot in our previous work as an example \cite{dong}, in this section, we construct the mathematical models regarding the body and arms to optimize their key design parameters and gaits. 

\subsection{Modeling of Crawling Locomotion}

\subsubsection{Force Analysis and Dynamic model}
\begin{figure}[t]
\centerline{\includegraphics[width=0.92\columnwidth]{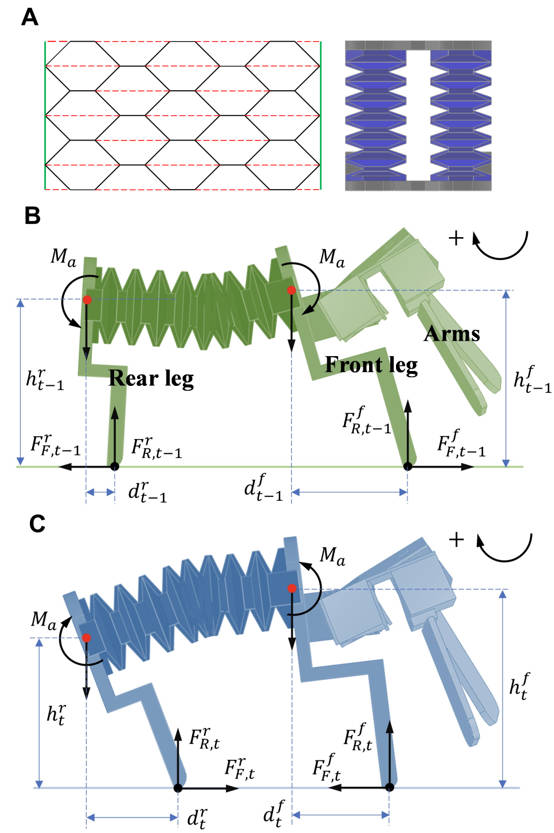}}
\vspace{-1mm}
\caption{\small{The origami sheet for the octagon-origami structure(B).  The black lines and the red dot lines represent mountain and valley folding patterns, respectively in (A). The force sketches of crawling captured at the time circle $t-1$ for the first half (B) and t for the second half(C). The sign ‘+’ represents the positive direction. 
$h$ denotes the height for the specified leg and $d$ is the horizontal distance between the reference point and the front foot for the given leg, respectively; $M_a$ is the equivalent actuation torque. The middle points (red points)of the ends of the octagon-origami structure are considered reference points.}}
\label{Fig:3}
\vspace{-7mm}
\end{figure}

The body of the proposed robots has an octagon-origami structure(as shown in Fig. \ref{Fig:3}-A) integrated with robot legs, which converts body deformation to linear motion to achieve a crawling gait. It mimics an inchworm with anisotropic friction, which has been adopted in other origami-inspired robots with similar gaits \cite{ref8}. The octagon-origami structure here keeps the same functionality and control schemes as \cite{ref31}.
The bending deformation of octagon-origami structures (robotic body) rises once the air pressure is exacted on these two octagon-origami structures for producing bending moments, which causes changes in interaction forces. Thus, the accurate actuation moment due to the air pressure is equivalent to two actuation moments at the ends of octagon-origami structures following the general beam theory, as seen in Fig. \ref{Fig:3}.

Here, the robot's forward motion is modeled into two phases. Assuming a generic time instant $t$, we define the rolling from the front leg as the first half of a cycle (at time step $t-1$) and that from the rear leg the second half (at time step $t$) \cite{ref20}. Then, a discrete dynamic equation is developed following Newton's Second Law, namely

\vspace{-1mm}
\begin{equation}
    \begin{matrix}
    \begin{aligned}
    m \ddot{x}_{t-1}&=F_{F,t-1}^{f}-F_{F,t-1}^{r}\\
    m \ddot{y}_{t-1}&=F_{R,t-1}^{f}+F_{R,t-1}^{r}-m \cdot g \\
    m \ddot{x}_{t}&=F_{F,t}^{f}-F_{F,t}^{r}\\
    m \ddot{y}_{t}&=F_{R,t}^{f}+F_{R,t}^{r}-m \cdot g 
    \end{aligned}
    \end{matrix}
\label{eq:5}
\end{equation}

\noindent With $F_{F}$=$\mu F_{R}$, where $m$ denotes the mass of the robot; $\mu$ is the  friction coefficient; $g$ is the gravitational acceleration;  $\ddot{x}$ and $\ddot{y}$ represent accelerations along the $x$-axis and $y$-axis, respectively. 

Combining the above equations, the ground reaction forces ($F_R$) during two phases of the cycle can be expressed as

\begin{equation}
    \begin{aligned}
    F_{R,t-1}^{f} &= \frac{m}{2} \left ( \frac{1}{\mu }\ddot{x}_{t-1}+ \ddot{y}_{t-1}+g  \right ) \\
    F_{R,t-1}^{r} &= \frac{m}{2} \left (-\frac{1}{\mu }\ddot{x}_{t-1}+ \ddot{y}_{t-1}+g  \right ) \\
    F_{R,t}^{f} &= \frac{m}{2} \left (-\frac{2}{\mu }\ddot{x}_{t}+ \ddot{y}_{t}+g  \right ) \\
    F_{R,t}^{r} &= \frac{m}{2} \left ( \frac{2}{\mu }\ddot{x}_{t}+ \ddot{y}_{t}+g  \right ) \\
    \end{aligned}
    \label{eq:6}
\end{equation}
Note, assuming a short time interval $\Delta T$, the accelerations are defined as
\begin{equation}
    \ddot{l} _{n} =\frac{l_{n+1}-2\cdot l_{n}+ l_{n-1}}{\bigtriangleup t^{2} } , l\in \left \{x, y \right \} 
    \label{eq:7}
\end{equation}

\begin{figure*}[t]
\centerline{\includegraphics[width=2\columnwidth]{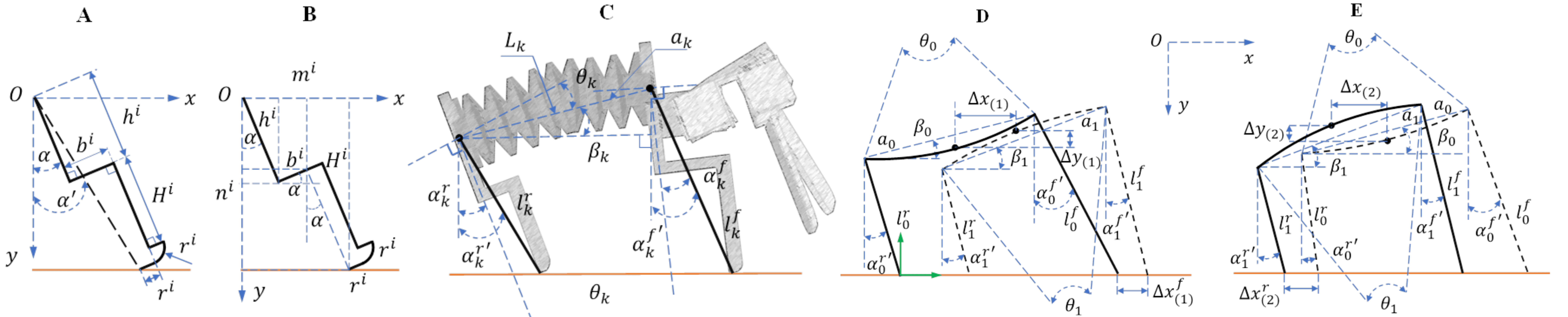}}
\vspace{-2.2mm}
\caption{\small{The geometrical and crawling sketches of the proposed robot. (A) The leg configuration with dimensions $h^i, H^i, b^i, r^i$, $i=l$ for the front leg and $i=r$ for the rear leg; (B), the projection lengths $m^i$ (along $x$-axis) and $n^i$ (along $y$-axis) directions; (C) The sketch of the proposed robot with geometrical parameters; (D-E) The status simplified sketches of the robot crawling in the first half and the second half. $\alpha_j^{{i}'}$ and $l_j^i$ symbolize the equivalent angle and the leg's length. $j$ being 0 or 1, denoting the first or second half. The subscripts 0 and 1 of the circumference angle $\theta$, the chord length $a$, and the chord inclination $\beta_k$ point to the sample time for a motion cycle. The robotic height can be mainly determined by the equivalent angle $\alpha_j^{{i}'}$ and leg length $l_j^f$  of the front leg.}}
\label{Fig:4}
\vspace{-4.5mm}
\end{figure*}

\subsubsection{Deformation Analysis of Crawling}
As illustrated in Fig.\ref{Fig:4}, the geometric deformation model is parameterized by the design parameters of the legs, $h^i, H^i, b^i, r^i$ ($i=l$ represents the front leg; $i=r$ denotes the rear leg.) and the variables, $\theta_k,\alpha_k^f,\alpha_k^r$ ($\theta_k$ is the circumference angle of the bending body;  $\alpha_k^f$ and $\alpha_k^r$  are the inclinations of the front and rear legs at the time circle $k$, respectively). We construct the model between design parameters and motion variables over the cycles of the first and second halves. The equivalent length $l_k^f$  and incline angle $\alpha_k^{{f}'}$ of the front leg by replacing the actual leg conFiguration for simplifying the calculation, with the projection lengths $m^f$ and $n^f$  along the $x$-axis and $y$-axis, respectively. 
\begin{equation}
    \resizebox{.88\hsize}{!}{$\begin{aligned}
    m^{f} &= h^{f} \cos\alpha _{k}^{f}  -\left (b^{f} -r^{f}   \right ) \sin \alpha _{k}^{f} +\left (H^{f} +r^{f}   \right ) \cos \alpha _{k}^{f}\\
    n^{f} &= h^{f} \sin\alpha _{k}^{f}  +\left (b^{f} -r^{f}   \right ) \cos \alpha _{k}^{f} +\left (H^{f} +r^{f}   \right ) \sin \alpha _{k}^{f}
    \end{aligned}$}
     \label{eq:8}
\end{equation}
\begin{equation}
    \begin{aligned}
    l_{k}^{f}&=\sqrt{m^{f^{2} }+n^{f^{2} } }  \\
    \alpha _{k}^{f^{\prime} } &= \arctan \left ( \frac{n^{f} }{m^{f} }  \right ) 
    \end{aligned}
    \label{eq:9}
\end{equation}
\noindent where the subscript $k$ represents the time instant. Similarly, the corresponding equivalent values of the rear leg are provided as
\begin{equation}
    \resizebox{.85\hsize}{!}{$
    \begin{aligned}
    m^{r} &= h^{r} \cos\alpha _{k}^{r}  -\left (b^{r} -r^{r}   \right ) \sin \alpha _{k}^{r} +\left (H^{r} +r^{r}   \right ) \cos \alpha _{k}^{r}\\
    n^{r} &= h^{r} \sin\alpha _{k}^{r}  +\left (b^{r} -r^{r}   \right ) \cos \alpha _{k}^{r} +\left (H^{r} +r^{r}   \right ) \sin \alpha _{k}^{r}
    \end{aligned}$}
    \label{eq:10}
\end{equation}
\begin{equation}
    \begin{aligned}
    l_{k}^{r}&=\sqrt{m^{r^{2} }+n^{r^{2} } }  \\
    \alpha _{k}^{r^{\prime} } &= \arctan \left ( \frac{n^{r} }{m^{r} }  \right ) 
    \end{aligned}
    \label{eq:11}
\end{equation} 
Fig.\ref{Fig:4} indicates that when combining Eq.(6) and Eq.(7), the mathematical model of the chord length $a_k$, the chord inclination $\beta_k$ and the circumference angle $\theta_k$ of the arc $L$ of the octagon-origami structure can be built based on the geometrical constraints as follows:
\begin{equation}
    \beta _{k} = \arcsin \left ( \frac{l_{k}^{f} \cos\alpha _{k}^{f^{\prime }}-l_{k}^{r} \cos\alpha _{k}^{r^{\prime } }  }{\alpha _{k} }  \right )
    \label{eq:12}
\end{equation}
\begin{equation}
    a_{k} =\frac{L}{\theta _{k} } \sin \theta _{k} 
    \label{eq:13}
\end{equation} 
\begin{equation}
    \begin{aligned}
    \theta _{k} = \alpha _{k}^{f} +\beta _{k} \\
    \theta _{k} = \alpha _{k}^{r} -\beta _{k} 
    \end{aligned}
    \label{eq:14}
\end{equation}
The equations above define the specific state of the robot. Combining them, we have all other variables that can be expressed via $\theta_k$, which means that the proposed robot has one degree of freedom for an octagon-origami structure.
$\Delta x_{(1)}, \Delta y_{(1)}$ during the first half and $\Delta x_{(2)}, \Delta y_{(2)}$ for the second are the displacements of the center of gravity along the $x$ and $y$  axes, respectively. The right superscript $i$ in $\Delta x_{(1)}^i$ is $f$ as the front leg or $r$ as the rear leg, as shown in Fig.\ref{Fig:4}(D, E). Note that when the radius of the arc formed from the octagon-origami structure is much larger than the chord of the arc, the arc length is almost equivalent to the arc chord. Therefore, the displacements of the center of the gravity are provided as

\vspace{-3mm}
\begin{equation}
\resizebox{.85\hsize}{!}{$\begin{matrix}
\Delta x_{(1)}+\frac{a_{1}}{2} \cdot \cos \beta_{1}=l_{1}^{f} \cdot \sin \alpha_{1}^{f^{\prime}}-l_{0}^{f} \cdot \sin \alpha_{0}^{f^{\prime}}+\frac{a_{0}}{2} \cdot \cos \beta_{0}+\Delta x_{(1)}^{f} \\
\Delta x_{(2)}+\frac{a_{1}}{2} \cdot \cos \beta_{1}=l_{0}^{r} \cdot \sin \alpha_{0}^{r^{\prime}}-l_{1}^{r} \cdot \sin \alpha_{1}^{r^{\prime}}+\frac{a_{0}}{2} \cdot \cos \beta_{0}+\Delta x_{(2)}^{r}
\end{matrix}$}
\end{equation} 
Due to rolling, the displacement of the leg is described as   
\begin{equation}
   \begin{array}{l}
\Delta x_{(1)}^{f}=r^{f} \cdot\left(\alpha_{1}^{f^{\prime}}-\alpha_{0}^{f^{\prime}}\right) \\
\Delta x_{(2)}^{r}=r^{r} \cdot\left(\alpha_{0}^{r^{\prime}}-\alpha_{1}^{r^{\prime}}\right) \\
\end{array}
\label{eq:16}
\end{equation}
\begin{equation}
\resizebox{.80\hsize}{!}{$\begin{matrix}
\begin{aligned}
    &\Delta y_{(1)} = \Delta y_{(2)}=\\
    &\frac{l_{1}^{f} \cos \alpha_{1}^{f^{\prime}}-l_{1}^{r} \cos \alpha_{1}^{r^{\prime}}}{2}-\frac{l_{0}^{f} \cos \alpha_{0}^{f^{\prime}}-l_{0}^{r} \cos \alpha_{0}^{r^{\prime}}}{2}
\end{aligned}
\end{matrix}$}
\label{eq:17}
\end{equation} 

 In terms of Eqs. (\ref{eq:8}-\ref{eq:17}), to arrive at a maximum displacement for each crawling circle,  the $\Delta x_{(1)}$ and $\Delta x_{(s)}$ should be enlarged. Moreover, $\Delta y_{(1)}$ and $\Delta y_{(2)}$ tend to be minimized, which enables the proposed robot to move stably.

 \subsection{Modeling of Swimming Locomotion}

\subsubsection{The Origami Twisted Tower}
\begin{figure*}[htbp]
\centerline{\includegraphics[width=1.85\columnwidth]{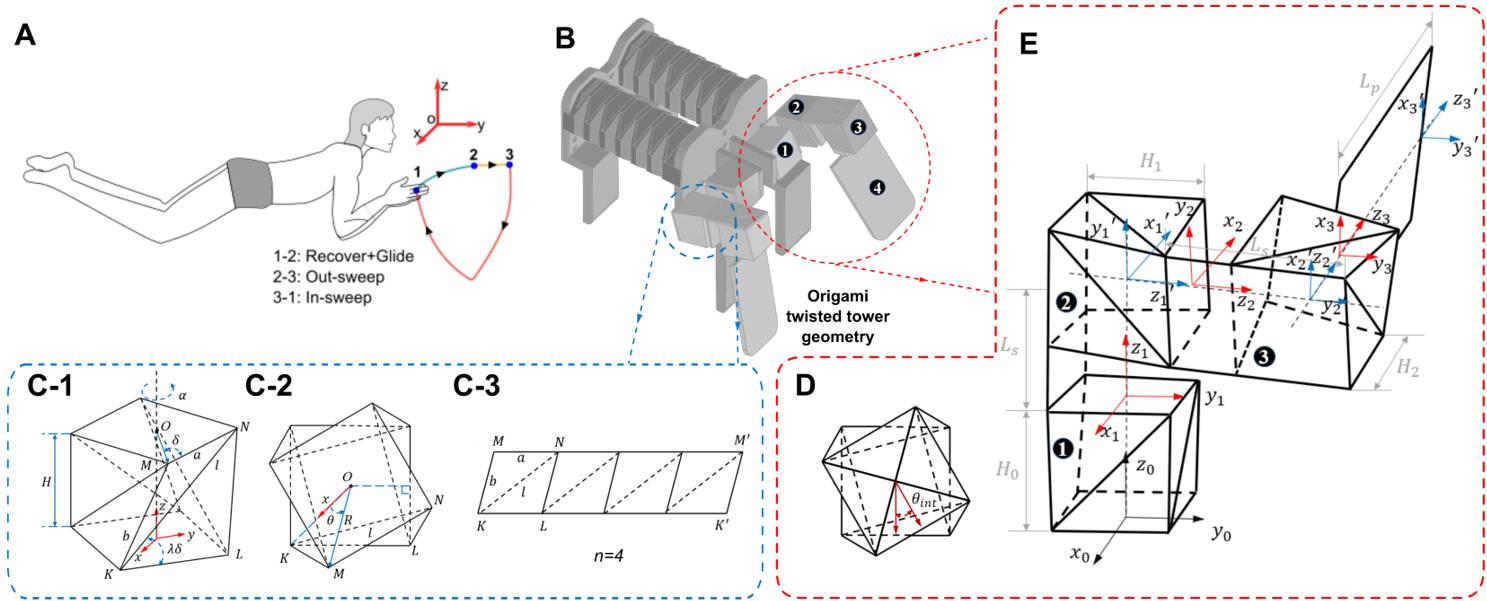}}
\vspace{-1mm}
\caption{\small{The Three Stages of Breaststroke locomotion(A), The sketch of the proposed robot(B); the front view(C-1), top-down view(C-2) and the corresponding crease pattern(C-3) of the origami twisted tower, where the solid black lines represent the mountain folds and the dashed black line denote the valley folds, respectively; the coordinate transformations for the first, second, third joint of the robotic arm(D); the top view of an origami joint(E). 1, 2, and 3 represent the lifting, outreaching, and rotating actions, and 4 indicates a flapping plate. $x_i,  y_i$ and $z_i$ denote the coordinate reference frame of the $i$-th joint.}}
\label{Fig:44}
\vspace{-4.5mm}
\end{figure*}

Inspired by human physiology, the proposed robot can realize human-like swimming locomotion based on three serially connected origami twist towers following the “generalized” Kresling pattern \cite{ref33}. Fig.\ref{Fig:44}-C presents the geometry of the origami twisted tower, where $K,L,N,M$ stand for the vertexes of the panel surface. The geometric parameters are defined: $a,b,l$ and $n$ represent the polygon side length, the panel side length, the diagonal valley-crease length, and the number of polygon sides, respectively. In addition, the angle $\delta$ is half the internal angle of the basal polygon, and $\lambda$, the angle ratio, is a transformation metric between open and closed states. With the rotation angle $\theta$ of the top polygon changing, the variation of vertexes in circumradius $R$ results in scaling the height $H$ of the origami structure, which can be expressed as:
\vspace{-1mm}
\begin{equation}
\vspace{-1.5mm}
    H=(b^2-2R^2(1-\cos\theta))^{\frac{1}{2}}
    \label{eq:18}
\end{equation} 
\vspace{-4.5mm}

The origami tower's dimensions are critical for design efficiency and ease of construction. Following Bhovad's work \cite{ref33}, 
we manually set $a=b=15$ mm, with a 95° angle, yielding an initial angle ($\theta_{int}\approx 7$°). 
To prevent interference between the rotating origami towers and the robot's body, we ensured adequate space by setting the separation ($L_s$) slightly larger than $a$, which is $16$ mm in this case, referring to Fig.\ref{Fig:44} and Fig.\ref{Fig:5} for clarity. Additionally, to ensure the robotic arms' effectiveness in pushing water during an outreach movement, the first joint is positioned as low as the design permits, maintaining a $10$ mm distance from the baseline. This careful sizing ensures the robot's arm movements are both functional and feasible.

\subsubsection{Forward Kinematics of 3-DoF Twisted Towers}
The forward kinematics of the 3-DoF origami twisted towers are established following the DH conventions, to illustrate the motion of the robotic arm. Considering that origami twisting tower output displacement and rotation simultaneously, we divide each tower into a composite of revolute joint and prismatic joint by adding a dummy coordinate. The red coordinate system in Fig.\ref{Fig:44}-E remains its orientation with the twisted tower's deformation, serving to effectively describe the tower's axial elongation at the P joints. Conversely, the blue coordinate system is attached to the tower's surface to articulate the rotational motion at the R joints. This dual-coordinate approach allows for a precise representation of the tower's complex motions. Table \ref{tbl:2} summarizes the D-H parameters.
Thus, the transformation matrix from joint $i$ to joint $i+1$ is:
\vspace{-1mm}
\begin{equation}
 \resizebox{.88\hsize}{!}{$T_{i(i+1)}=\left[\begin{array}{cccc}
\cos \theta^{i} & -\cos \alpha_{i} \sin \theta^{i} & \sin \alpha_{i} \sin \theta^{i} & a_{i} \cos \theta^{i} \\
\sin \theta^{i} & \cos \alpha_{i} \cos \theta^{i} & -\sin \alpha_{i} \cos \theta^{i} & a_{i} \sin \theta^{i} \\
0 & \sin \alpha_{i} & \cos \alpha_{i} & d^{i} \\
0 & 0 & 0 & 1
\end{array}\right]$}
\label{eq:19}
\end{equation} 
 
As shown in Fig.\ref{Fig:44}-D, we have
\begin{equation}
\centering
\begin{matrix}
\theta_i^t = \theta _{int} + \theta_i\\
d_0 = H_0 +L_s, d_1 = H_1 + L_s, d_2 = H_2 +L_p
\end{matrix}
\label{eq:20}
\end{equation}

\noindent where subscript $i \in \left \{ 1, 2, 3 \right \} $ denotes the $i$-th link of the robotic arm; $\theta_i$ and $H_i$ represent the rotation angle and the height with such a rotation angle; $L_s$ and $L_p$ denote the supporting link length and the padding link length, as shown in Fig.\ref{Fig:44}-E; $\theta_i^t$ and $d_i$ indicate the total rotation angle and the total length, correspondingly. Substituting DH parameters into the Eq.\ref{eq:19}, the final $T$ can be expressed with Eq.\ref{eq:21}:
\begin{align}
\centering
\resizebox{1\hsize}{!}{$
\begin{matrix}
&T=    
&\left[\begin{array}{cccc}
c \theta_{0}^{t} c \theta_{1}^{t} c \theta_{2}^{t}-s \theta_{0}^{t} s \theta_{2}^{t} & -c \theta_{0}^{t} c \theta_{1}^{t} s \theta_{2}^{t}-s \theta_{0}^{t} c \theta_{2}^{t} & -c \theta_{0}^{t} s \theta_{1}^{t} & -d_{2} c \theta_{0}^{t} s \theta_{1}^{t}-0.5 a \cdot s \theta_{0}^{t}+d_{1} s \theta_{0}^{t} \\
s \theta_{0}^{t} c \theta_{1}^{t} c \theta_{2}^{t}+c \theta_{0}^{t} s \theta_{2}^{t} & -s \theta_{0}^{t} c \theta_{1}^{t} s \theta_{2}^{t}+c \theta_{0}^{t} c \theta_{2}^{t} & -s \theta_{0}^{t} s \theta_{1}^{t} & -d_{2} s \theta_{0}^{t} s \theta_{1}^{t}+0.5 a \cdot c \theta_{0}^{t}-d_{1} c \theta_{0}^{t} \\
s \theta_{1}^{t} c \theta_{2}^{t} & -s \theta_{1}^{t} s \theta_{2}^{t} & c \theta_{1}^{t} & d_{2} c \theta_{1}^{t}-0.5 a+d_{0} \\
0 & 0 & 0 & 1
\end{array}\right]
\end{matrix}$}
\label{eq:21}
\end{align}

\noindent where $c$ and $s$ represent the $cos$ and $sin$, respectively. Thus, the reachable trajectories of the arm can be obtained.

\begin{table}[htbp]
\caption{The standard D-H parameters of the proposed robotic arm(Separation of coupled variables)}
\vspace{-1mm}
\centering
\setlength{\tabcolsep}{4.5mm}{
\scalebox{0.90}{\begin{tabular}{cccccc}
\hline
$i$    & $\alpha_i$              & $a_i$    & $d_i$  & $\theta_i$ & offset  \\ \hline
1      & 0                  & 0  & $\widehat{H_1}$ & 0   & 0         \\
${1}'$ & $90^{\circ}$                  & 0             & $L_s$  & $\widehat{\theta^1}$ & $180^{\circ}$\\
2      & 0               & $-90^{\circ}$ & $\widehat{H_2}/2$ & 0  & 0          \\
${2}'$ & $90^{\circ}$               & 0 & $L_s-a/2$ & $\widehat{\theta^2}$  & $90^{\circ}$          \\
3      & 0 & 0             & $\widehat{H_3}$  & 0 & 0\\
${3}'$ & 0 & 0             & $L_p$  & $\widehat{\theta^3}$ & $-80^{\circ}$\\
\hline
\end{tabular}}}
\label{tbl:2}
\vspace{-2mm}
\end{table}

\subsubsection{Problem Description on Gait Optimization}
The forward efficiency of the proposed robot relies not only on its structural design but heavily on an optimized gait. Drag-based swimming, a common aquatic locomotion mode, involves alternating protraction and retraction of paddle-like limbs, with the power stroke exerting more force than the recovery stroke \cite{drag-based-swimming}.  Modulating the effective surface area of appendages has been proven an effective solution. Ideally, Paddles should maximize their area during the power stroke and minimize resistance during the non-propulsive phase by reorienting into a stream-lined shape \cite{Swimming_back}. 
Taking inspiration from human breaststroke swimming, an algorithm is proposed to provide hints for optimized swimming gaits and can be generalized to various multi-joint articulated arms consisting of soft actuation elements.

Technically in the breaststroke, arm stroke is divided into four phases, namely Recovery, glide, out-sweep, and in-sweep, as seen in Fig. \ref{Fig:Swimming}-A. These phases involve constantly changing hand path width, depth, and pose to achieve improved efficiency \cite{Swimming_factors}. Since the Out-sweep does not contribute to propulsion and thrust is primarily generated in the In-sweep \cite{Outsweep}, we simplify the analysis to focus on the propulsive (In-sweep) and drag strokes (Out-sweep and Recovery), aiming for maximum thrust and minimum drag, respectively.  Here, two key points are paid attention to, namely the start and end points of the in-sweep phase. Specifically, the start is the point in the workspace that spreads out the furthest, with the maximum $x$ value, and the end corresponds to the point with the minimum $y$ value.

Referring to the blade element theory \cite{Swimming_area}, the thrust force is defined as:
\begin{equation}
    F_{d}=\frac{1}{2} C_{d} \rho A_{p} V^{2}
\label{eq: thrust}
\end{equation}
where $\rho$, $C_{d}$, $A_{p}$, and $V$ are fluid density, drag coefficient, projected area, and relative velocity, respectively. 
To quantify the efficiency along a given path, the forward strength is determined as the integral of thrust force in its forward direction (y+), generated with a unit surface area.  At this point, the problem becomes a trajectory optimization problem, where the optimization objective is to maximize and minimize the volume of water pushed corresponding to the trajectory.

\subsubsection{Human-like Swimming Gaits Optimization}
To solve this problem, we model the question as a graph theoretic problem in the discredited space. Considering a uniform swimming motion, where $V$ is constant, the thrust is linearly proportional to the projected area with the appendages, as illustrated in Eq. \ref{eq: thrust}. A rasterized joint-space point cloud is obtained by isometric sampling of the joint angles, and then forward kinematics is used to obtain the corresponding pose, including position and orientation. The
specific projected area is derived by projecting the normal of the end plane with the y-axis. Thus, for each node, we calculate its incremental thrust strength with its neighboring nodes in the joint space, which is the multiplication of the difference in the y-axis with the averaged projected areas. The goal is to find the path that maximizes and minimizes the volume of thrust strength along the path. $A*$ algorithm is used for path planning, which balances efficiency and optimality. Specifically, for the thrust stroke, the heuristic function is defined as the largest projected area multiplied by the difference in y-coordinate. This allows the heuristic function to overestimate the thrust strength and ensure optimality so as to provide a path with maximum thrust strength. Conversely, for the drag strokes, the goal is to minimize the total dragging that impedes the motion. Thus, the heuristic function is set as the multiplication of the smallest projected area with the difference in y-coordinate values between the current point and the target point. Namely,

\begin{equation}
\begin{matrix}
    H_{thrust}(P_n) = \mathop{max}\limits_{P_n \in P}(A_{p}(Pn))|y_n - y_{goal}| \\
    H_{drag}(P_n) = \mathop{min}\limits_{P_n \in P}(A_{p}(Pn))|y_n - y_{goal}|
\end{matrix}
\label{eq: heuristic}
\end{equation}

\begin{figure}[t]
\centerline{\includegraphics[width=0.99\columnwidth]{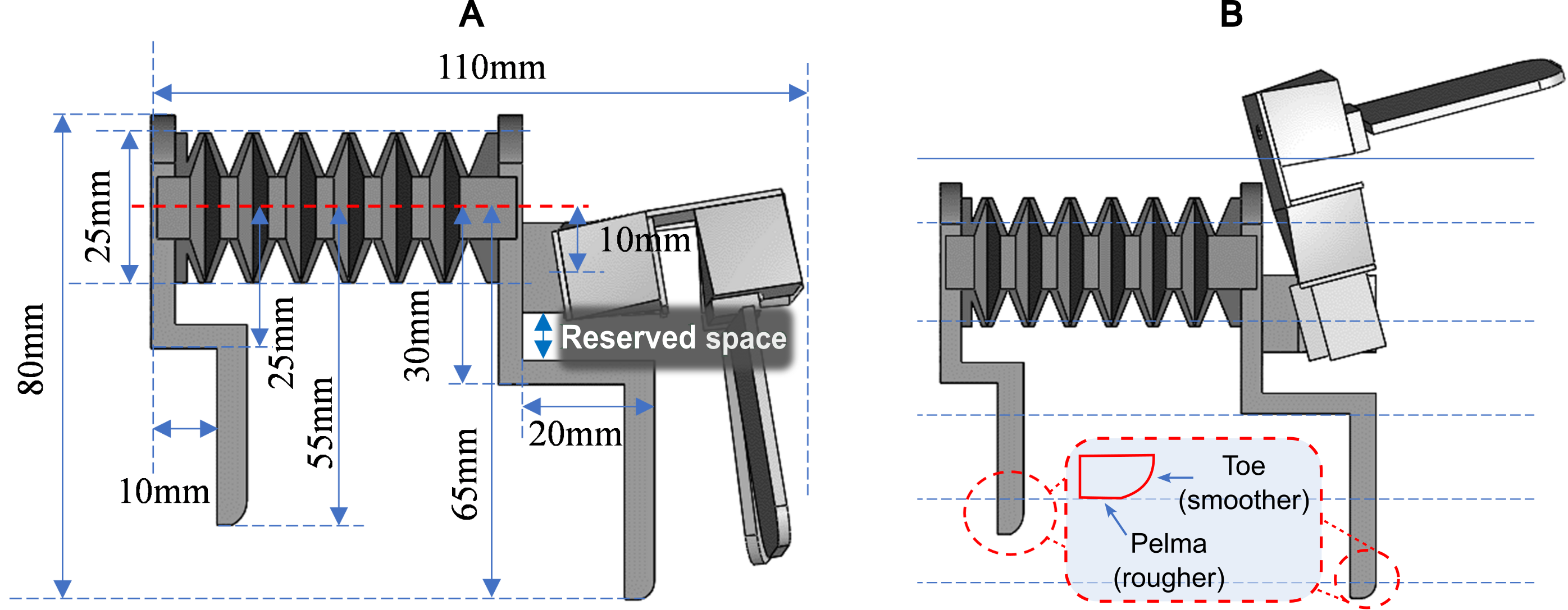}}
\vspace{-1.5mm}
\caption{\small{The dimensions of the proposed robot(A) and the first joint of the proposed robot lifts arms in the water(B). The red dashed line indicates the dimension baseline(A). The blue dash lines represent the water(B). A blown-up view of the robotic foot is provided, which has two frictional surfaces.}}
\label{Fig:5}
\vspace{-4mm}
\end{figure}

\section{Results and Analysis}
In this part, we analytically evaluate the proposed robot's multiple-locomotion capabilities based on the proposed modeling, including crawling analysis and swimming trajectory optimization. 
\subsection{Crawling Locomotion Friction Optimization}
The built dynamic force model is applied to determine the suitable design for achieving the optimal movement speed and stability of the robot within the constraints of geometrical ranges (see Fig.\ref{Fig:5}), with a maximum height of 85mm and length of 110mm.

\begin{figure*}[htbp]
\vspace{0mm}
\centerline{\includegraphics[width=2\columnwidth]{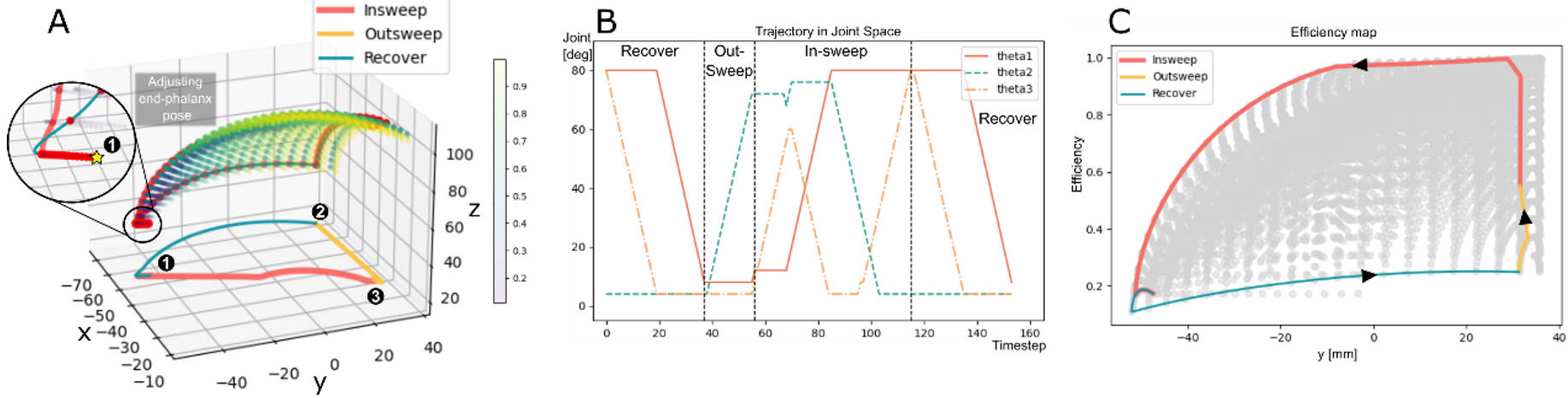}}
\vspace{-1.5mm}
\caption{\small{(A) The workspace of the 3-DOF origami-based arm and
the obtained optimal swimming arm gaits, including thrust stroke
(in-sweep) and drag stroke (out-sweep and recovery); (B) The joints trajectory of the obtained gaits. (C) The relationship of nodal projected area (efficiency) along the y-axis}}
\label{Fig:Swimming}
\vspace{-5.5mm}
\end{figure*}

\subsubsection{The robotic foot design based on the simulation}
\begin{figure}[t]
\vspace{0mm}
\centerline{\includegraphics[width=0.95\columnwidth]{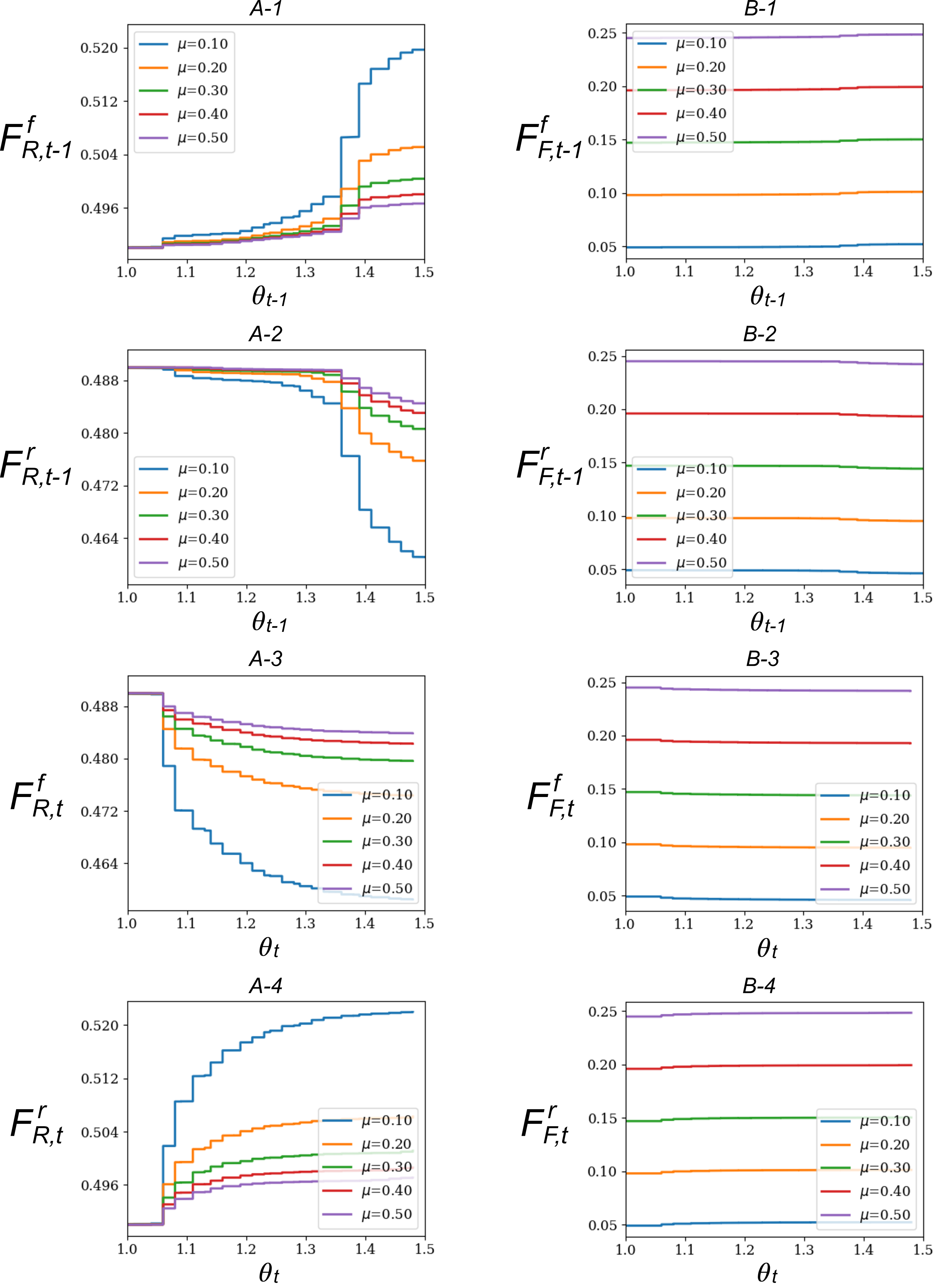}}
\vspace{-1.5mm}
\caption{\small{The ground reaction forces $F_{R,t-1}^f$, $F_{R,t-1}^r$ for the first half cycle(A-1, A-2) and $F_{R,t}^f$, $F_{R,t}^r$ for the second half cycle (A-3, A-4) in terms of the circle angles $\theta_{t-1}$, $\theta_t$ of arcs actuated by the control system at the time circles $t-1$ and $t$, and (B) shows the friction, respectively.}}
\label{Fig:6}
\vspace{-5.5mm}
\end{figure}

The dynamic model is used for determining the surface materials to determine the proposed legs' surface materials by evaluating the ground reaction forces during one cycle. The corresponding parameters have been set, namely $g=9.8(m/s^2)$, and $m=0.1(kg)$. For simplicity, we consider five discrete values of friction coefficients ($\mu \in \left \{ 0.1k\mid k=1, ..., 5 \right \} $) to investigate their influence on crawling efficiency. Fig.\ref{Fig:6}(A-1, A-2) and (A-3, A-4) depict the magnitudes of the ground reaction forces experienced by the front and rear feet at sequential time circles $t-1$ and $t$, respectively, derived by Eqs. (\ref{eq:6}-\ref{eq:17}). Ground reaction forces are computed by multiplying directly with the friction coefficient to ascertain the contact frictional force between the foot and the ground, shown in Fig. \ref{Fig:3}. As evidenced in Fig. \ref{Fig:3} and Fig. \ref{Fig:5}-B, during the time circle $t-1$, the toes of the rear foot and the pelma surface of the front foot are concurrently in contact with the ground. To optimize forward locomotion at this juncture, maximum the distance traveled in a time cycle(Eqs. (\ref{eq:16},\ref{eq:17})), it is crucial to immobilize the front foot while propelling the rear foot forward by leveraging the frictional interaction with the ground. The converse is applicable at time circle $t$. 

Consequently, a superior frictional coefficient is attributed to the plantar surfaces of both the front and rear feet compared to the toe sections. This differential in frictional attributes facilitates maximal forward progression during the time circle $t-1$ and minimizes resistive effects at time circle $t$.

\subsection{Swimming Locomotion Trajectory Optimization}

\subsubsection{The optimization results and simulation}
Fig.\ref{Fig:Swimming}-A presents the workspace of the 3-DOF origami-based arm and obtained optimal swimming arm gaits, including thrust stroke (in-sweep) and drag stroke (out-sweep and recovery), with the color corresponding to the size of the unit projected area at the given point. It can be seen that the proposed two-state planned trajectory aligns well with the arm trajectory of human breaststroke swimming (Fig. \ref{Fig:44}-A).

Due to the mechanical characteristics and the spatial alignment of the three origami twisted towers, the resultant workspace is a piece of a thick spherical shell. It should be noted that at the beginning of the recovery stroke, the planned trajectory passes through the workspace point cloud,  with the rapid decrease of $\theta_3$ as seen in Fig. \ref{Fig:Swimming}-B. This can be ascribed to the large difference in the projected area at the interior and exterior of the workspace in Cartesian space. The conduct of the trajectory search in the joint space rather than Cartesian space provides trajectories that exhibit superior continuity and smoothness, avoiding over-quick deformation that is not feasible for fragile soft structures. With systematical gait planning, the proposed robot can rapidly adjust the orientation of the end at the junction of the two stages, shifting the projected area from a maximum case to the minimum state in the shortest distance. Thus we alleviate the influence of dragging in the swimming cycle, demonstrating the effectiveness of our proposed algorithm. 
Fig.\ref{Fig:Swimming}-C presents the relationship of nodal projected area (efficiency) along the y-axis in the workspace. The area enclosed by a specific curve is equivalent to the thrust/drag produced by that gait.

\begin{figure}[t]
\centerline{\includegraphics[width=0.92\columnwidth]{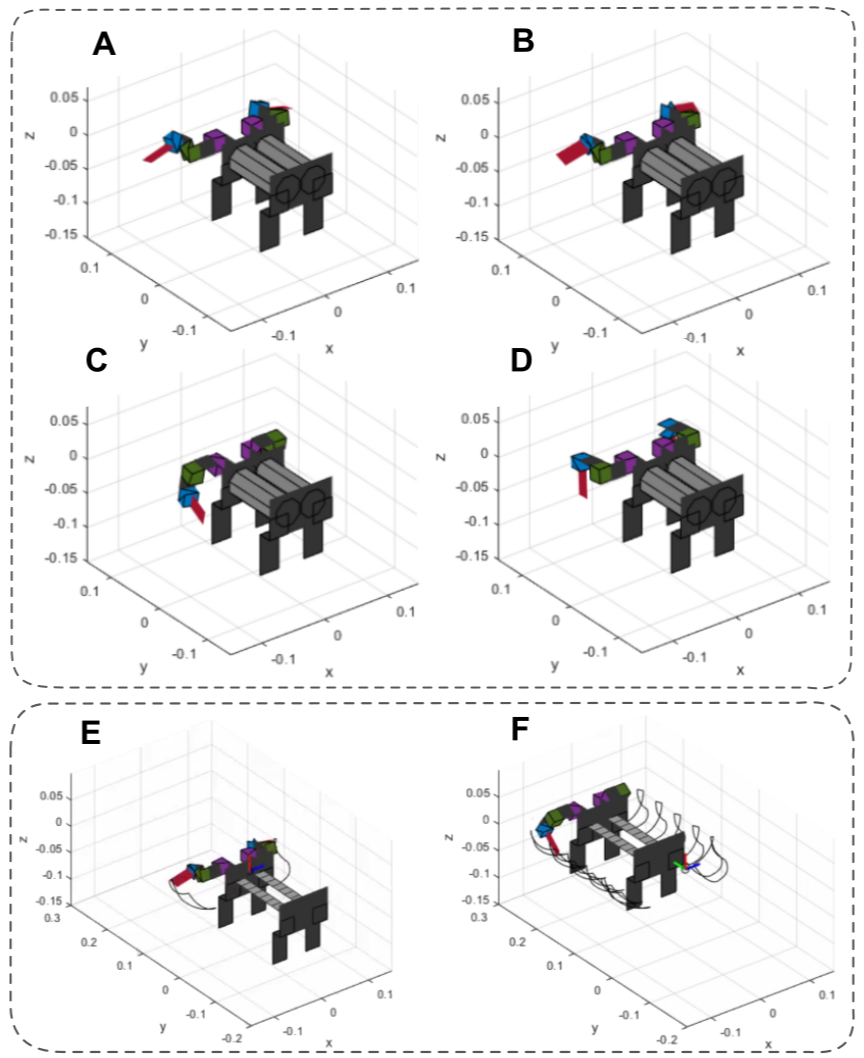}}
\vspace{-2mm}
\caption{\small{The snapshots of the robot swimming locomotion, (A)end of out-sweep; (B) beginning of in-sweep; (C) beginning of recover; (D) beginning of out-sweep; (E) and (F) demonstrate the trajectory of the end of the arm as the robot swims forward.}}
\vspace{1mm}
\label{Fig:simulation}
\vspace{-4mm}
\end{figure}

To better understand the posture of the robot while swimming, we built a simulation model of the robot in Matlab. Each side of the twisted tower was accurately modeled to obtain a more realistic swimming state. Using the three-degree-of-freedom robotic arm joint rotation angle we obtained through the A* trajectory search algorithm, we simulated the posture of the robot's breaststroke throughout the cycle. From Fig.\ref{Fig:simulation}-A to Fig.\ref{Fig:simulation}-C is the entire insweep stage, which is responsible for efficiently giving the robot forward force; Fig.\ref{Fig:simulation}-C to Fig.\ref{Fig:simulation}-D, Fig.\ref{Fig:simulation}-D to Fig.\ref{Fig:simulation}-B are the recovery stage respectively. and outweep stages, our algorithm ensures that in these two stages, the force hindering the robot's progress is minimal. What is worth noting is the change in posture from Fig.\ref{Fig:simulation}-B to Fig.\ref{Fig:simulation}-A. From the end of the previous cycle to the beginning of the next cycle, the third joint will rotate independently, converting the end of the arm from a low-resistance water-cutting posture to a high-resistance The water-pushing posture is more in line with bionics. Fig.\ref{Fig:simulation}-E, F show the trajectory left by the robot during its advancement under multiple action cycles. Compared with our previous research\cite{dong}, the swimming efficiency has been greatly improved, and it is more like a breaststroke posture.

\subsection{Discussion}

\begin{figure}[t]
\centerline{\includegraphics[width=1.05\columnwidth]{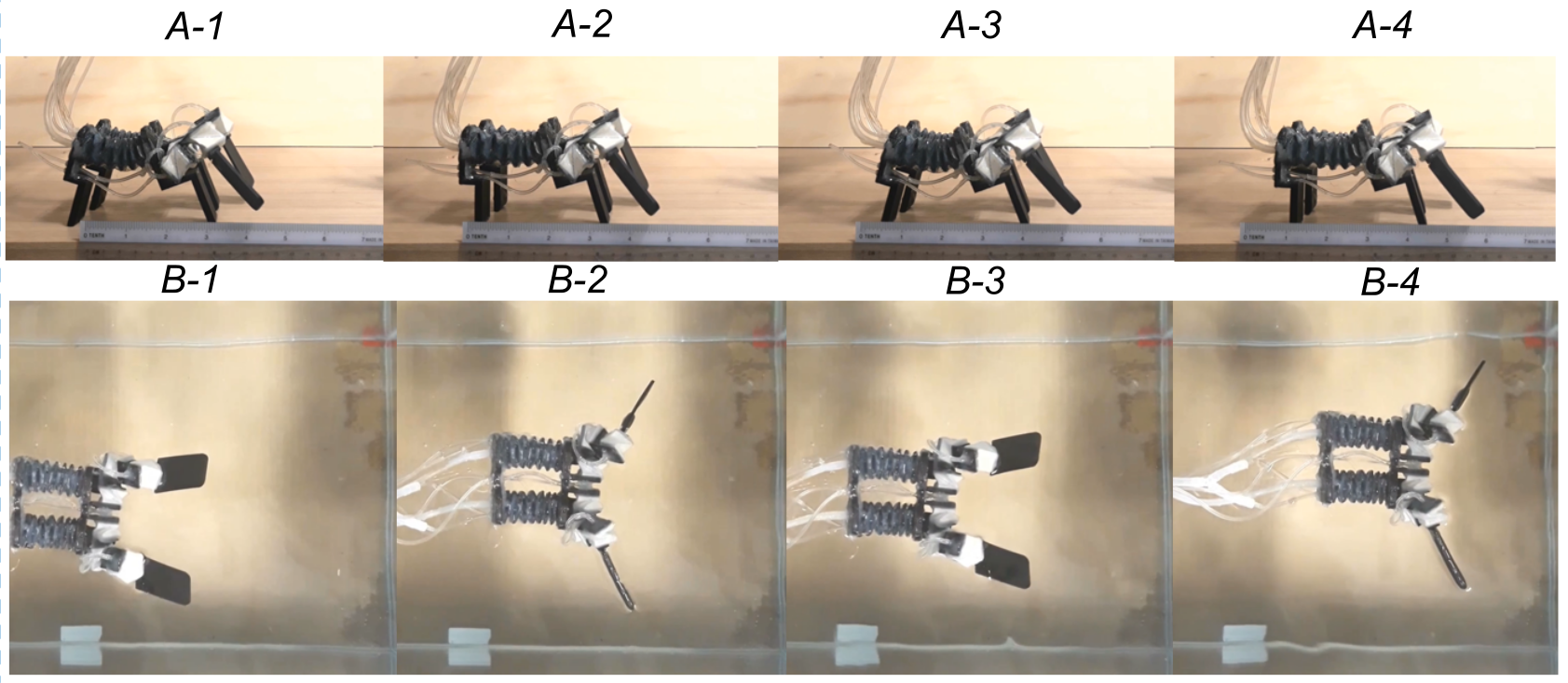}}
\vspace{-2mm}
\caption{\small{The snapshots of the robot experiment in physical scenarios\cite{dong} (A) demonstrate swimming locomotion, and (B) for crawling locomotion }}
\vspace{1mm}
\label{Fig:experiment}
\vspace{-4mm}
\end{figure}

The physical performance of the multiple-locomotion capabilities based on the proposed modeling has been verified and presented in our previous work \cite{dong}, as seen in Fig. \ref{Fig:experiment}. With the optimized parameters and gaits, the proposed robot is able to perform stable crawling and swimming locomotion. 

We have provided the corresponding comparisons with other works. The parameter optimization method we proposed is based on simplified beam theory with linear deformation, while Soft materials such as rubber, silicone, etc. often have complex nonlinear relationships. Compared with the topology-based methods with finite elements analysis (FEA) illustrated in \cite{topology}, we only obtain the qualitative relationship among the design parameters. Besides, in contrast to the dynamic modeling proposed in \cite{tro}, the proposed robot has shortcomings in the capability of underwater locomotion and velocity control for buoyancy and 
turning. The current method plans the gaits in the position map by assuming a constant forward velocity, neglecting the positive influence of velocity on the dragging force produced by water. However, this simple strategy can be easily generalized to a wider range of soft robots to improve their efficiency. Meanwhile, although the body is designed for crawling, swimming efficiency can be further improved through arm-leg coordination by considering the contribution and impedance of the legs.

\section{CONCLUSIONS}
In this work, we constructed mathematical models to provide the theoretical foundation for the design optimization and control of soft origami robots on crawling and swimming locomotion. By combining the periodic dynamic model of the robots with the design parameters and motion variables, we qualitatively reveal the influence of the surface frictional coefficient on crawling motion, and provide hints for crawling-based robot design. Besides, a swimming kinematics model, together with a heuristic-based algorithm, has been developed to optimize the swimming trajectory with the least effort. Both simulations and experiments were carried out to illustrate the effectiveness of the proposed modeling and gaits control strategies, which allow the proposed robot to successfully imitate an inchworm that crawls on the ground and humans that swim in the water, respectively. Future research will be focused on improving the capability of underwater locomotion for such robots. We will continue exploring the dynamic modeling of the proposed robot to further optimize the design and research on arm-leg coordination for improved performance. 
\vspace{1mm}
\AtNextBibliography{\small}
\printbibliography

\end{document}